\documentclass[runningheads]{llncs}
\usepackage{graphicx}
\usepackage{tikz}
\usepackage{comment}
\usepackage{amsmath,amssymb} % define this before the line numbering.
\usepackage{color}
\usepackage{booktabs}

\usepackage{hyperref} % for url
\hypersetup{hidelinks,
	colorlinks=true,
	allcolors=black,
	pdfstartview=Fit,
	breaklinks=true}
	
\newcommand{\etal}{{\emph{et al.}}}

\usepackage[accsupp]{axessibility}  % Improves PDF readability for those with disabilities.

\usepackage{float}
\begin{document}
% \linenumbers
\pagestyle{headings}
\mainmatter
\def\ECCVSubNumber{4505}  % Insert your submission number here

\title{MoFaNeRF: Morphable Facial Neural \\ Radiance Field} % Replace with your title

%\begin{comment}
%\titlerunning{ECCV-22 submission ID \ECCVSubNumber} 
%\authorrunning{ECCV-22 submission ID \ECCVSubNumber} 
%\author{Anonymous ECCV submission}
%\institute{Paper ID \ECCVSubNumber}
%\end{comment}
%******************

%\begin{comment}
\titlerunning{MoFaNeRF: Morphable Facial Neural Radiance Field}
\author{Yiyu Zhuang\inst{*} \and
Hao Zhu\inst{*} \and
Xusen Sun\and
Xun Cao$^\dag$}
\authorrunning{Y. Zhuang et al.}
\institute{Nanjing University, Nanjing, China\\
\email{\{yiyu.zhuang, xusensun\}@smail.nju.edu.cn \{zhuhaoese, caoxun\}@nju.edu.cn}}
%\end{comment}
%******************
\maketitle

\begin{abstract}
We propose a parametric model that maps free-view images into a vector space of coded facial shape, expression and appearance with a neural radiance field, namely Morphable Facial NeRF. Specifically, MoFaNeRF takes the coded facial shape, expression and appearance along with space coordinate and view direction as input to an MLP, and outputs the radiance of the space point for photo-realistic image synthesis. Compared with conventional 3D morphable models (3DMM), MoFaNeRF shows superiority in directly synthesizing photo-realistic facial details even for eyes, mouths, and beards. Also, continuous face morphing can be easily achieved by interpolating the input shape, expression and appearance codes.
By introducing identity-specific modulation and texture encoder, our model synthesizes accurate photometric details and shows strong representation ability. Our model shows strong ability on multiple applications including image-based fitting, random generation, face rigging, face editing, and novel view synthesis. Experiments show that our method achieves higher representation ability than previous parametric models, and achieves competitive performance in several applications. To the best of our knowledge, our work is the first facial parametric model built upon a neural radiance field that can be used in fitting, generation and manipulation. The code and data is available at \underline{\url{https://github.com/zhuhao-nju/mofanerf}}.

\end{abstract}

\let\thefootnote\relax\footnotetext{$^*$ These authors contributed equally to this work.}
\let\thefootnote\relax\footnotetext{ $^\dag$ Xun Cao is the corresponding author. }

\section{Introduction}
\label{sec:intro}

\begin{figure}[t]
    \centering
    \includegraphics[width=1.0\linewidth]{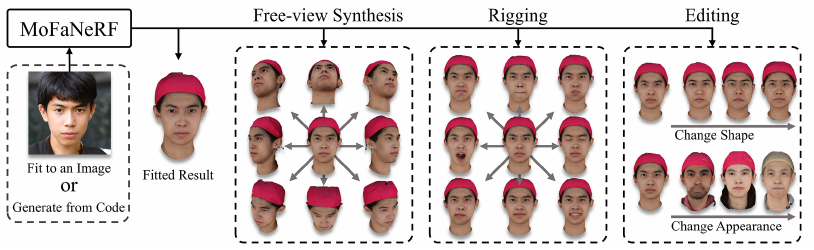}
    \vspace{-0.25in}
    \caption{We propose MoFaNeRF, which is a parametric model that can synthesize free-view images by fitting to a single image or generating from a random code. The synthesized face is \emph{morphable} that can be rigged to a certain expression and be edited to a certain shape and appearance.
    }
    \label{fig:title}
    \vspace{-0.2in}
\end{figure}

% introduce 3DMM and its problem
Modeling 3D face is a key problem to solve face-related vision tasks such as 3D face reconstruction, reenactment, parsing, and digital human. The 3D morphable model (3DMM)\cite{blanz1999morphable} has long been the key solution to this problem, which is a parametric model transforming the shape and texture of the faces into a vector space representation.
3DMMs are powerful in representing various shapes and appearances, but require a sophisticated rendering pipeline to produce photo-realistic images. Besides, 3DMMs struggled to model non-Lambertian objects like pupils and beards.
Recently, the neural radiance field (NeRF)\cite{mildenhall2020nerf} was proposed to represent the shapes and appearances of a static scene using an implicit function, which shows superiority in the task of photo-realistic free-view synthesis. The most recent progress shows that the modified NeRF can model a dynamic face\cite{gafni2021dynamic,wang2021learning,park2021hypernerf,park2020deformable}, or generate diversified 3D-aware images\cite{schwarz2020graf,chan2021pi,gu2021stylenerf}. However, there is still no method to enable NeRF with the abilities of single-view fitting, controllable generation, face rigging and editing at the same time. In summary, conventional 3DMMs are powerful in representing large-scale editable 3D faces but lack the ability of photo-realistic rendering, while NeRFs are the opposite. %The advantages of 3DMM and NeRF complement each other.

% Motivation and challenge
To combine the best of 3DMM and NeRF, we aim at creating a facial parametric model based on the neural radiance field to have the powerful representation ability as well as excellent free-view rendering performance. However, achieving such a goal is non-trivial.
The challenges come from two aspects: firstly, how to memorize and parse the very large-scale face database using a neural radiance field; secondly, how to effectively disentangle the parameters (e.g. shape, appearance, expression), which are important to support very valuable applications like face rigging and editing.

% Our contributions
To address these challenges, we propose the Morphable Facial NeRF (MoFaNeRF) that maps free-view images into a vector space of coded facial identity, expression, and appearance using a neural radiance field.  Our model is trained on two large-scale 3D face datasets, FaceScape\cite{yang2020facescape,zhu2021facescape} and HeadSpcae\cite{Dai2019} separately. FaceScape contains $359$ available faces with $20$ expressions each, and HeadSpace contains $1004$ faces in the neutral expression. The training strategy is elaborately designed to disentangle the shape, appearance and expression in the parametric space.
The identity-specific modulation and texture encoder are proposed to maximize the representation ability of the neural network. Compared to traditional 3DMMs, MoFaNeRF shows superiority in synthesizing photo-realistic images even for pupils, mouth, and beards which can not be modeled well by 3D mesh models. Furthermore, we also propose the methods to use our model to achieve image-based fitting, random face generation, face rigging, face editing, and view extrapolation. Our contributions can be summarized as follows:

\begin{itemize}

    \item To the best of our knowledge, we propose the first parametric model that maps free-view facial images into a vector space using a neural radiance field and is free from the traditional 3D morphable model.
    
    \item The neural network for parametric mapping is elaborately designed to maximize the solution space to represent diverse identities and expressions. The disentangled parameters of shape, appearance and expression can be interpolated to achieve a continuous and morphable facial synthesis. 
    
    \item We present to use our model for multiple applications like image-based fitting, view extrapolation, face editing, and face rigging. Our model achieves competitive performance compared to state-of-the-art methods.
    
\end{itemize}

\section{Related Work}
\label{sec:related}

As our work is a parametric model based on neural radiance field, we will review the related work of 3D morphable model and neural radiance field respectively.

% 3DMM related

\noindent\textbf{3D Morphable Model.} 3DMM is a statistical model which transforms the shape and texture of the faces into a vector space representation\cite{blanz1999morphable}. By optimizing and editing parameters, 3DMMs can be used in multiple applications like 3D face reconstruction\cite{xiao2022detailed}, alignment\cite{jourabloo2016large}, animation\cite{zhang2021flow}, etc.  We recommend referring to the recent survey\cite{egger20203d} for a comprehensive review of 3DMM.
To build a 3DMM, traditional approaches first capture a large number of 3D facial meshes, then align them into a uniform topology representation, and finally process them with principal component analysis algorithm\cite{yang2020facescape,zhu2021facescape,vlasic2005face,cao2013facewarehouse,li2017learning,jiang2019disentangled}.  The parameter of the 3DMM can be further disengaged into multiple dimensions like identity, expression, appearance, and poses.
In recent years, several works tried to enhance the representation power of 3DMM by using a non-linear mapping\cite{bagautdinov2018modeling,tewari2018self,tran2019learning,tran2018nonlinear,cheng2019meshgan,tran2019towards}, which is more powerful in representing detailed shape and appearance than transitional linear mapping. However, they still suffer from the mesh representation which is hard to model fine geometry of pupils, eyelashes and hairs. Besides, traditional 3DMMs require sophisticated rendering pipelines to render photo-realistic images. By contrast, our model doesn't explicitly generate shape but directly synthesizes photo-realistic free-view images even for pupils, inner-mouth and beards.

Very recently, Yenamandra~\etal~\cite{yenamandra2021i3dmm} proposed to build the 3DMM with an implicit function representing facial shape and appearance. They used a neural network to learn a signed distance field(SDF) of 64 faces, which can model the whole head with hair. Similarly, our model is also formulated as an implicit function but very different from SDF. SDF still models shape while our method focuses on view synthesis and releases constraints of the shape, outperforming SDF in rendering performance by a large margin.

% NeRF related
\noindent\textbf{Neural Radiance Field.} NeRF\cite{mildenhall2020nerf} was proposed to model the object or scene with an impressive performance in free-view synthesis. NeRF synthesizes novel views by optimizing an underlying continuous volumetric scene function that is learned from multi-view images. 

As the original NeRF is designed only for a static scene, many efforts have been devoted to reconstructing deformable objects.
Aiming at the human face many methods\cite{gafni2021dynamic,wang2021learning,park2021hypernerf} modeled the motion of a single human head with a designed conditional neural radiance field, extending NeRF to handle dynamic scenes from monocular or multi-view videos.
Aiming at human body, several methods have been proposed by introducing human parametric model (e.g. SMPL)\cite{noguchi2021neural,chen2021animatable,liu2021neural,peng2021neural} or skeleton\cite{peng2021animatable} as prior to build NeRF for human body. For a wide range of dynamic scenarios, Park \etal~\cite{park2020deformable} proposed to augment NeRF by optimizing an additional continuous volumetric deformation field, while Pumarola \etal~\cite{pumarola2021d} optimized an underlying deformable volumetric function.
Another group of works \cite{schwarz2020graf,chan2021pi,gu2021stylenerf} turned NeRF into a generative model that is trained or conditioned on certain priors, which achieves 3D-aware images synthesis from a collection of unposed 2D images. 
To reduce the image amount for training, many works~\cite{yu2021pixelnerf,wang2021ibrnet,raj2021pixel,Gao20arxiv_pNeRF} trained the model across multiple scenes to learn a scene prior, which achieved reasonable novel view synthesis from a sparse set of views.

Different from previous NeRFs, our method is the first parametric model for facial neural radiance field trained on a large-scale multi-view face dataset. Our model supports multiple applications including random face generation, image-based fitting and facial editing, which is unavailable for previous NeRFs.

%==================================================
\section{Morphable Facial NeRF}
\label{sec:method}

\begin{figure}[t]
    \centering
    \includegraphics[width=1.0\linewidth, trim=0 0.2in 0 0]{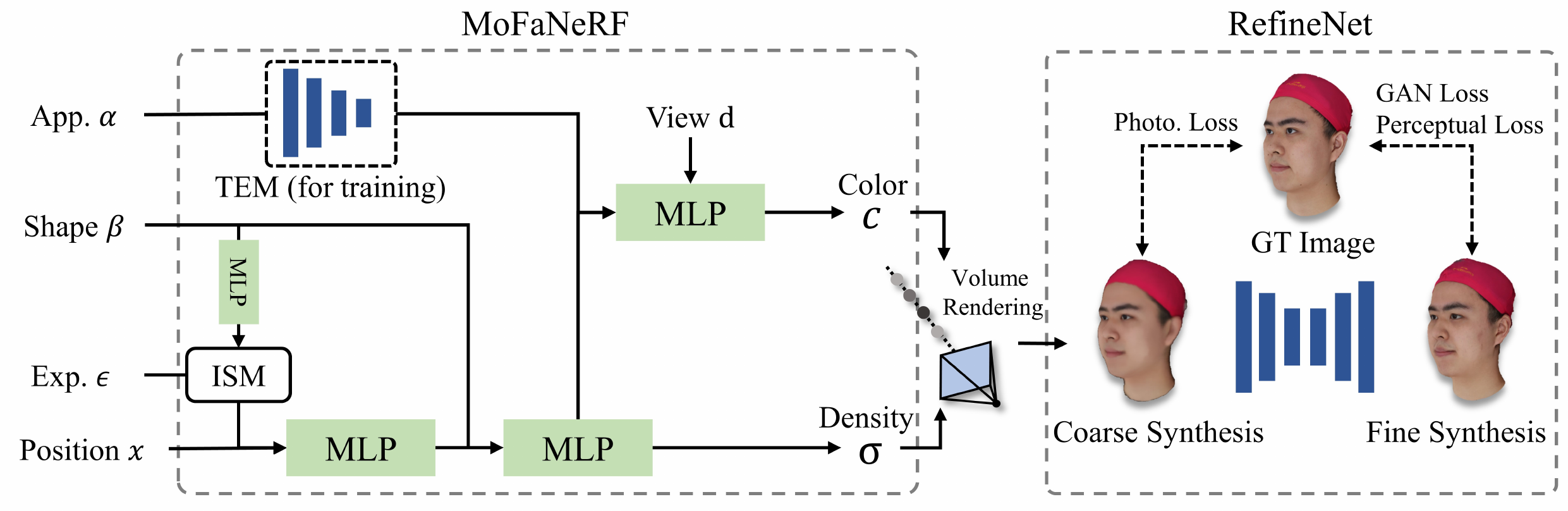}
    \caption{MoFaNeRF takes appearance code $\alpha$, shape code $\beta$, expression code $\epsilon$, position code $\textbf{x}$ and view direction $\textbf{d}$ as input, synthesizing a coarse result which is then refined by a RefineNet. As shown in the right bottom corner, MoFaNeRF can be used in generating (synthesize free-view images given parameters) or fitting (optimize for parameters given a single image).}
    \label{fig:network}
    \vspace{-0.2cm}
\end{figure}

Morphable facial NeRF is a parametric model that maps free-view facial portraits into a continuous morphable parametric space, which is formulated as:

\begin{equation}
    \mathcal{M}: (\mathbf{x,d,\beta,\alpha,\epsilon}) \to \{\mathbf{c,\sigma}\},
\end{equation}

where $\mathbf{x}$ is the 3D position of a sample point; $\mathbf{d}$ is the viewing direction consisting of pitch and yaw angles; $\beta, \alpha, \epsilon$ are the parameters denoting facial shape, appearance, and expression respectively; $\mathbf{c}$ and $\sigma$ are the RGB color and the density used to represent the neural radiance field. In the next, we will explain $\mathbf{x, d, c,\sigma}$ that are referred from NeRF in Section~\ref{sec:nerf}, then introduce $\beta, \alpha, \epsilon$ in Section~\ref{sec:param}. The network design is illustrated in Section~\ref{sec:network} and the training details are explained in Section~\ref{sec:train}.

\subsection{Neural Radiance Field}
\label{sec:nerf}

As defined in NeRF\cite{mildenhall2020nerf}, the radiance field is represented as volumetric density $\sigma$ and color $\mathbf{c}=(R,G,B)$. An MLP is used to predict $\sigma$ and $\mathbf{c}$ from a 3D point $\mathbf{x}=(x,y,z)$ and viewing direction $\mathbf{d}=(\theta,\phi)$. Position encoding is introduced to transform the continuous inputs $\mathbf{x}$ and $\mathbf{d}$ into a high-dimensional space, which is also used in our model.
The field of $\sigma$ and $\mathbf{c}$ can be rendered to images using a differentiable volume rendering module. For a pixel in the posed image, a ray $\mathbf{r}$ is cast through the neural volume field from the ray origin $\mathbf{o}$ along the ray direction $\mathbf{d}$ according to the camera parameters, which is formulated as $\mathbf{r}(z)=\mathbf{o}+z\mathbf{d}$.
Through sampling points along this ray, and accumulating the sampled density $\sigma(\cdot)$ and RGB values $\mathbf{c}(\cdot)$ computed by $\mathcal{F}$, the final output color $\mathbf{C(r)}$ of this pixel can be evaluated by:
\begin{equation}
    \mathbf{C(r)}\!=\!\int_{z_{n}}^{z_{f}} T(z) \sigma( \mathbf{r}(z)) \mathbf{c}(\mathbf{r}(z),\mathbf{d})dz\,, 
     \:\textup{where}\:T(z)\!=\!\exp\!\left(-\int_{z_{n}}^{z}\sigma (\mathbf{r}(s))\mathrm{ds}\right) .
\end{equation}
 $T(t)$ is defined as the accumulated transmittance along the ray from $z_n$ to $z$, where $z_n$ and $z_f$ are near and far bounds.
Through the rendered color, a photometric loss can be applied to supervise the training of the MLP.

\subsection{Parametric Mapping}
\label{sec:param}

Our model is conditioned on the parameters to represent the identity and facial expression $\epsilon$, and the identity is further divided into shape $\beta$ and appearance $\alpha$. Initially, we consider integrating $\beta$ and $\alpha$ into a single identity code, however, we find it is hard for an MLP to memorize the huge amount of appearance information. Therefore, we propose to decouple the identity into shape and appearance. These parameters need to be disentangled to support valuable applications like face rigging and editing.

\noindent\textbf{Shape parameter $\beta$} represents the 3D shape of the face that is only related to the identity of the subject, like the geometry and position of the nose, eyes, mouth and overall face. A straightforward idea is to use one-hot encoding to parameterize $\beta$, while we find it suffers from redundant parameters because the similarity of large-amount faces is repeatedly expressed in one-hot code.  Instead, we adopt the identity parameters of the bilinear model of FaceScape\cite{yang2020facescape} as shape parameter, which is the PCA factors of the 3D mesh for each subject. The numerical variation of the identity parameter reflects the similarity between face shapes, which makes the solution space of facial shapes more efficient. 

\noindent\textbf{Appearance parameter $\alpha$} reflects photometric features like the colors of skin, lips, and pupils. Some fine-grained features are also reflected by appearance parameters, such as beard and eyelashes. Considering that the UV texture provided by FaceScape dataset is the ideal carrier to convey the appearance in a spatial-aligned UV space, we propose to encode the UV texture maps into $\alpha$ for training.  The texture encoding module (TEM) is proposed to transfer the coded appearance information into the MLP, which is a CNN based encoder network. TEM is only used in the training phase, and we find it significantly improves the quality of synthesized images. We consider the reason is that the appearance details are well disentangled from shape and spatial-aligned, which relieves the burden of memorizing appearances for the MLP.

\noindent\textbf{Expression parameter $\epsilon$} is corresponding to the motions caused by facial expressions. Previous methods\cite{park2021hypernerf,Tretschk20arxiv_NR-NeRF} try to model the dynamic face by adding a warping vector to the position code $\textbf{x}$, namely deformable volume.
However, our experiments show that the deformable volume doesn't work in our task where too many subjects are involved in a single model. More importantly, our training data are not videos but images with discrete 20 expressions, which makes it even harder to learn a continuous warping field.
By contrast, we find directly concatenating expression parameters with the position code as \cite{li2021dynerf,gafni2021dynamic} causes fewer artifacts, and our identity-specific modulation (detailed in Section~\ref{sec:network}) further enhances the representation ability of expression. We are surprised to find that MLP without a warping module can still synthesize continuous and plausible interpolation for large-scale motions. We believe this is the inherent advantage of the neural radiance field over 2D-based synthesis methods.  

\subsection{Network Design}
\label{sec:network}

As shown in Figure~\ref{fig:network}, the backbone of MoFaNeRF mainly consists of MLPs, identity-specific modulation (ISM) module and texture encoding module(TEM). These networks transform the parameters $\alpha, \beta, \epsilon$, position code $\textbf{x}$ and viewing direction $\textbf{d}$ into the color $\textbf{c}$ and density $\sigma$. The predicted colors are then synthesized from $\textbf{c}$ and $\sigma$ through volume rendering. Considering that the appearance code $\alpha$ is only related to the color $c$, it is only fed into the color decoder.  
The expression code $\epsilon$ is concatenated to the position code after the identity-specific modulation, as it mainly reflects the motions that are intuitively modulated by shape $\beta$. The RefineNet takes the coarse image predicted by MoFaNeRF as input and synthesizes a refined face. The results presented in this paper are the refined results by default. The additional texture encoding module (TEM) is used only in the training phase, which consists of $7$ convolution layers and $5$ full connected layers . The detailed parameters of our network are shown in the supplementary.

To represent a large-scale multi-view face database, the capacity of the network needs to be improved by increasing the number of layers in MLP and the number of nodes in each hidden layer. The generated images indeed gets improved after enlarging the model size, but is still blurry and contains artifacts in the expressions with large motions. 
To further improve the performance, we present the identity-specific modulation and RefineNet. 

\noindent\textbf{Identity-specific modulation (ISM).} Intuitively, facial expressions of different individuals differ from each other as individuals have their unique expression idiosyncrasies. However, we observed that the MLPs erase most of these unique characteristics after the disentanglement, homogenizing the expressions from different subjects.
Motivated by AdaIN\cite{karras2019style,karras2020analyzing}, we consider the unique expression of individuals as a modulation relationship between $\beta$ and $\epsilon$, which can be formulated as:
\begin{equation}
    \epsilon'=M_s(\beta)\cdot\epsilon+M_b(\beta),
\end{equation}
where $\epsilon'$ is the updated value to the expression code, $M_s$ and $M_b$ are the shallow MLPs to transform $\beta$ into an identity-specific code to adjust $\epsilon$. Both $M_s$ and $M_b$ output tensors with the same length as $\epsilon$. Our experiments show that ISM improves the representation ability of the network especially for various expressions. 

\noindent\textbf{RefineNet.}
We propose to take advantage generative adversial networks to further improve the synthesis of the facial details. We use Pix2PixHD\cite{wang2018high} as the backbone of RefineNet, which refine the results of MoFaNeRF with GAN loss\cite{goodfellow2014generative} and perceptual loss\cite{johnson2016perceptual}. The input of RefineNet is the coarse image rendered by MoFaNeRF, and the output is a refined image with high-frequency details. We find that RefineNet significantly improves details and realism with less impact on identity-consistency. The influence of RefineNet on identity-consistency are validated in the ablation study in Section~\ref{sec:disen_eval}.

\subsection{Training}
\label{sec:train}

\noindent\textbf{Data preparation.} We use $7180$ models released by FaceScape\cite{yang2020facescape} and $1004$ models released by HeadSpace\cite{Dai2019} to train two models respectively.  In FaceScape, the models are captured from $359$ different subjects with $20$ expressions each. For FaceScape, we randomly select $300$ subjects ($6000$ scans) as training data, leaving $59$ subjects ($1180$ scans) for testing. For HeadSpace, we randomly select $904$ subjects as training data, leaving $100$ subjects for testing. As HeadSpace only consists of a single expression for each subjects, the expression input part of the network to train HeadSpace data is removed. All these models are aligned in a canonical space, and the area below the shoulder is removed. We render $120$ images in different views for each subjects. The details about the rendering setting are shown in the supplementary.

\noindent\textbf{Landmark-based sampling.} In the training phase, the frequency of ray-sampling is modified besides the uniform sampling to make the network focus on the facial region. Specifically, we dectect $64$ 2D key-points of the mouth, nose, eyes, and eyebrows, and the inverse-projecting rays are sampled around each key-point based on a Gaussian distribution. The standard deviation of the Gaussian distribution is set to $0.025$ of the image size all our experiments. The uniform sampling and the landmark-based sampling are combined with the ratio of 2:3.

\noindent\textbf{Loss function.}
The loss function to train MoFaNeRF is formulated as:

\begin{equation}
    L=\sum\limits _{\mathbf{r}\in \mathcal{R}} \left [ \left \| \hat{C}_c(\mathbf{r}) - C(\mathbf{r})  \right \|^2_2 +
\left \| \hat{C}_f(\mathbf{r}) - C(\mathbf{r}) \right \|^2_2  \right],
\end{equation}
where $\mathcal{R}$ is the set of rays in each batch, $C(\mathbf{r})$ is the ground-truth color, $\hat{C}_c(\mathbf{r})$ and $\hat{C}_f(\mathbf{r})$ are the colors predicted by coarse volume and fine volume along ray $\mathbf{r}$ respectively. It is worth noting that the expression and appearance parameters are updated according to the back-propagated gradient in the training, while the shape parameters remain unchanged.  We firstly train the network of MoFaNeRF, then keep the model fixed and train the RefineNet. The RefineNet is trained with the loss function following Pix2PixHD\cite{wang2018high}, which is the combination of GAN loss\cite{goodfellow2014generative} and perceptual loss\cite{dosovitskiy2016generating,gatys2016image,johnson2016perceptual}. The implementation details can be found in the supplementary material.

\begin{figure}[tb]
    \centering
    \includegraphics[width=1\linewidth, trim=0 0.0cm 0 0]{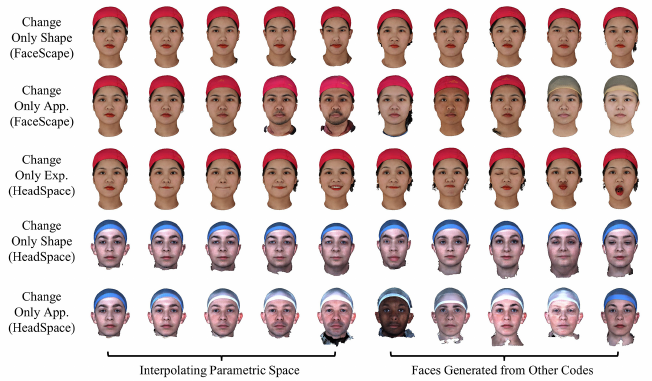}
    \caption{Our model is able to synthesize diverse appearance, shape and expressions, while these three dimensions are well disentangled. The interpolation in the parametric space shows that the face can morph smoothly between two parameters.
    }
    \label{fig:morph}
    \vspace{-0.1cm}
\end{figure}

%==================================================
\subsection{Application}
\label{sec:app}

In addition to directly generating faces from a certain or random vector, MoFaNeRF can also used in image-based fitting, face rigging and editing.

\begin{figure}[t]
    \centering
    \includegraphics[width=1.0\linewidth,trim=0 0in 0 0]{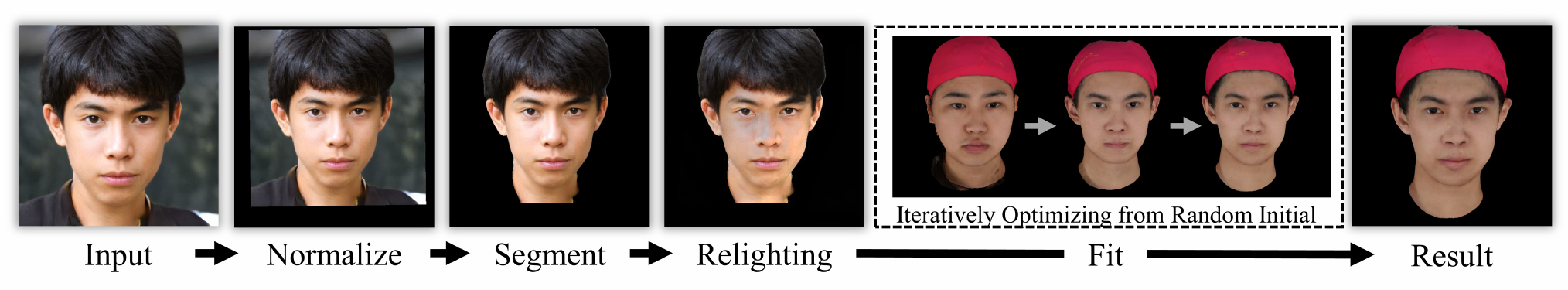}
    \vspace{-0.3in}
    \caption{The pipeline for fitting our model to a single image.}
    \label{fig:fit_pipeline}
    \vspace{-0.1in}
\end{figure}

\noindent\textbf{Image-based fitting.} As shown in Figure~\ref{fig:fit_pipeline}, we propose to fit our model to an input image. Firstly, we normalize the image to the canonical image with an affine transformation. Specifically, we first extract the 2D landmarks $L_t$ from the target image with \cite{kazemi2014one} , then align $L_t$ to the predefined 3D landmarks $L_c$ of the canonical face by solving:
\begin{equation}
    \textbf{d, s} = \arg\min \left \|(\Pi (L_c, \textbf{d})) \cdot \textbf{s} - L_t \right \|_2,
\end{equation}
where $\textbf{d}$ is the view direction, $\textbf{s}$ is the scale. $\Pi(L_c, $\textbf{d}$)$ is the function to project 3D points to the 2D image plane according to the view direction $\textbf{d}$. The scale $\textbf{s}$ is applied to the target image, and $\textbf{d}$ is used in the fitting and remains constant. Then we use EHANet~\cite{luo2020ehanet,CelebAMask-HQ} to segment the background out, and normalize the lighting with the relighting method~\cite{zhou2019deep}. In practice, we find it important to eliminate the influence of light because our model cannot model complex lighting well.

After the pre-processing, we can optimize for $\beta, \alpha, \epsilon$ through the network. Specifically, $\beta$ and $\alpha$ are randomly initialized by Gaussian distribution, and $\epsilon$ is initialized with the learned value from the training. Then we freeze the pre-trained network weights and optimize $\alpha, \beta, \epsilon$ through the network by minimizing only the MSE loss function between the predicted color and the target color. Only points around landmarks are sampled in fitting.

\noindent\textbf{Face rigging and editing.} The generated or fitted face can be rigged by interpolating in expression dimension with controllable view-point. The expression vector can be obtained by fitting to a video or manually set. Currently, we only use the basic $20$ expressions provided by FaceScape to generate simple expression-changing animation. By improving the rigging of the face to higher dimensions\cite{li2010example}, our model has the potential to perform more complex expressions. The rigged results are shown in Figure~\ref{fig:title}, Figure~\ref{fig:morph} and the supplementary materials.

The generated or fitted face can be edited by manipulating the shape and appearance code. As explained in Section~\ref{sec:param}, shape coder refers to the shape of the face, the geometry and position of the nose, eyes, and mouth; while appearance refers to the color of skin, lips, pupils, and fine-grained features like beard and eyelashes. These features can be replaced from face A to face B by simply replacing the shape or appearance code, as shown in Figure~\ref{fig:title}. Our model supports manually editing by painting texture map, then using TEM to generate appearance code for a generation. However, we find only large-scale features of the edited content in the texture map will take effect, like skin color and beard, while small-scale features like moles won't be transferred to the synthesized face. We also demonstrate that the face can morph smoothly by interpolating in the vector space, as shown in Figure~\ref{fig:morph}.

\section{Experiment}
\label{sec:exp}

\begin{table}[t]
\centering
\caption{Quantitative evaluation of representation ability.}
\begin{tabular}{@{}lccc@{}}
\toprule
Model    & PSNR(dB)$\uparrow$ & SSIM$^*$$\uparrow$ & LPIPS$^*$$\downarrow$ \\ \midrule
FaceScape\cite{yang2020facescape}      & 27.96$\pm$1.34    & 0.932$\pm$0.012    & 0.069$\pm$0.009   \\
FaceScape{$^*$}\cite{yang2020facescape}      &  27.07$\pm$1.46   & 0.933$\pm$0.011    & 0.080$\pm$0.014   \\
i3DMM\cite{yenamandra2021i3dmm}          & 24.45$\pm$1.58    & 0.904$\pm$0.014    & 0.112$\pm$0.015   \\
MoFaNeRF & \textbf{31.49$\pm$1.75}    & \textbf{0.951$\pm$0.010}    & 0.061$\pm$0.011 \\
MoFaNeRF-fine & 30.17$\pm$1.71    & 0.935$\pm$0.013    & \textbf{0.034$\pm$0.007}\\ \bottomrule
\end{tabular}
\label{tab:represent}
\end{table}

\begin{figure}[t]
    \centering
    \includegraphics[width=0.7\linewidth]{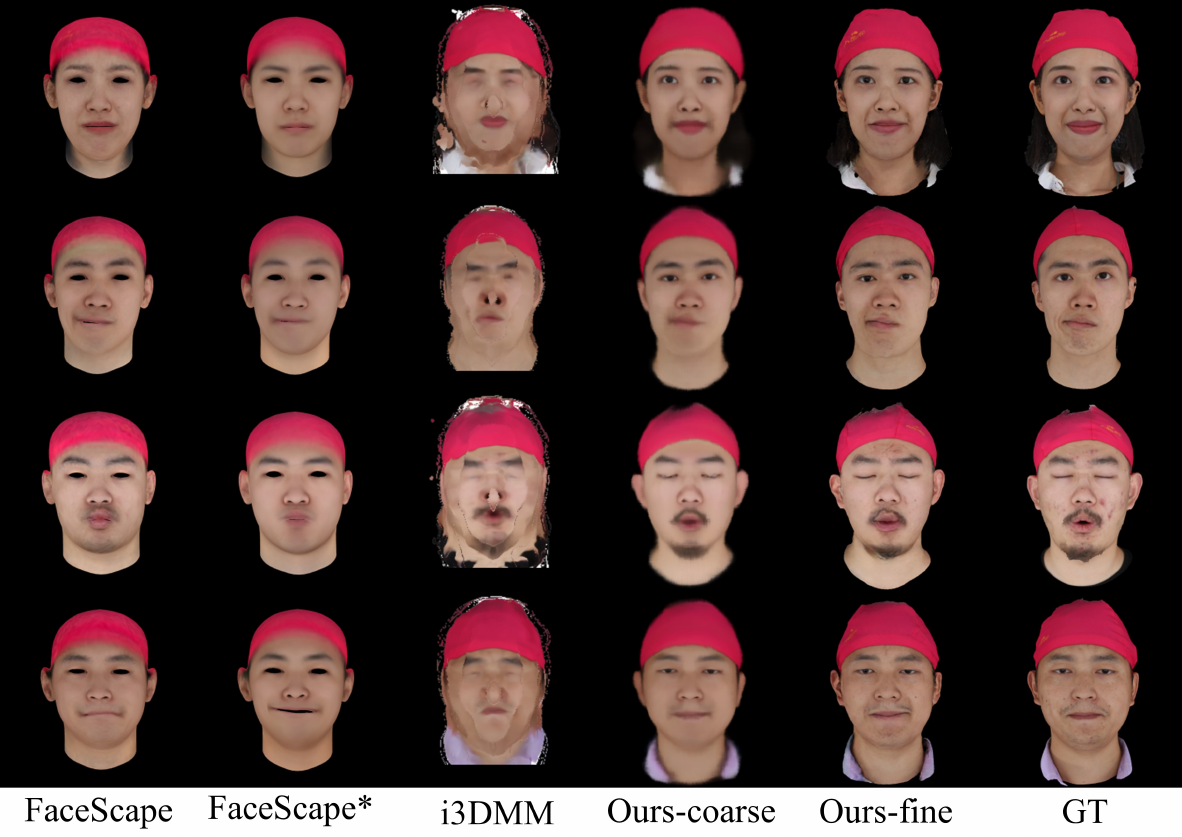}
    \vspace{-0.1in}
    \caption{Visual comparison of representation ability. Facescape$^*$ is the smaller version with comparable model size to our model($\approx120$M).}
    \vspace{-0.1in}
    \label{fig:represent}
\end{figure}

We firstly compare our model with previous parametric models in representation ability, then show the effectiveness of the parameter disentanglement and the network design in the ablation study. Finally, we evaluate the performance of MoFaNeRF in single-view image-based fitting, view extrapolation, and face manipulation.

\subsection{Comparison of Representation Ability}

We compare the representation ability of our MoFaNeRF with two SOTA facial parametric models - FaceScape bilinear model\cite{yang2020facescape} and i3DMM\cite{yenamandra2021i3dmm}.  FaceScape is the traditional 3DMM that applies PCA to 3D triangle mesh, while i3DMM is the learning-based 3DMM that represents shape via SDF. Both models are trained on FaceScape dataset as described in Section~\ref{sec:train}.  The default generated number of parameters for FaceScape is very large($\approx630$M), so to be fair, we also generated a model with a similar number of parameters to our model($\approx120$M), labeled as FaceScape$^*$. 
PSNR\cite{PSNRvsSSIM}, SSIM\cite{wang2004image} and LPIPS\cite{zhang2018unreasonable} are used to measure the objective, structural, and perceptual similarity respectively.  The better performance in similarity between the generated face image and ground truth indicates better representation ability.

From the visual comparison in Figure~\ref{fig:represent}, we can see that the FaceScape bilinear model doesn't model pupils and inner mouth, as it is hard to capture accurate 3D geometry for these regions. The rendered texture is blurry due to the misalignment in the registration phase and the limited representation ability of the linear PCA model.  i3DMM is able to synthesize the complete head, but the rendering result is also blurry. We observed that the performance of i3DMM trained on our training set has degraded to some extent, and we think it is because our data amount is much larger than theirs ($10$ times larger), which makes the task more challenging.  By contrast, our model yields the clearest rendering result, which is also verified in quantitative comparison shown in Table~\ref{tab:represent}. The refinement improves the LPIPS but decrease PSNR and SSIM, we believe this it is because the GAN loss and perceptual loss focus on hallucinate plausible details but is less faithful to the original image.

\begin{table}[t]
\centering
\caption{Validation of identity consistency.}
\begin{tabular}{@{}lccc@{}}
\toprule
Setting  & before RefineNet & after RefineNet & ground-truth \\ \midrule
changing view   & 0.687$\pm$0.027 & 0.707$\pm$0.028 & 0.569$\pm$0.048 \\
changing exp, view  & 0.703$\pm$0.023 & 0.720$\pm$0.025 & 0.633$\pm$0.029\\
 \bottomrule
\end{tabular}

\label{tab:ablation_view}
\vspace{0.1cm}
\end{table}

\subsection{Disentanglement Evaluation}
\label{sec:disen_eval}

We show the synthesis results of different parameters in the right side of Figure~\ref{fig:morph} to demonstrate that shape, appearance and expression are well disentangled, and shown the interpolation of different attributes in the left side of Figure~\ref{fig:morph} to demonstrate that the face can morph continuously.

We further validate identity-consistency among different views and different expressions. The distance in facial identity feature space (DFID) defined in FaceNet~\cite{schroff2015facenet} is used to measure how well the identity is preserved. Following the standard in FaceNet, two facial images with DFID $\le 1.1$ are judged to be the same person. We use a subset of our training set for this experiment, containing $10$ subjects with $10$ expressions each. We evaluate DFID between the ground truth and the fitted face rendered in $50$ views with heading angle in $0\sim90^{\circ}$. As reported in Table~\ref{tab:ablation_view}, the DFID scores after RefineNet are slightly increased, but are still comparable to the DFID scores of the ground-truth images. The DFID scores of both changing view and changing view and expression are much lower than $1.1$, which demonstrates that the RefineNet doesn't cause severe identity inconstancy across rendering view-points.

\begin{table}[t]
\centering
\caption{Ablation study.}
\begin{tabular}{@{}lccc@{}}
\toprule
Label  & PSNR(dB)$\uparrow$ & SSIM$\uparrow$ & LPIPS$\downarrow$ \\ \midrule
(a.1)One-hot expression code $\epsilon$\  &25.59$\pm$2.25    &0.888$\pm$0.025    &0.184$\pm$0.039   \\
(a.2)PCA expression code $\epsilon$\ &25.79$\pm$2.25    &0.886$\pm$0.025    &0.187$\pm$0.039   \\
(a.3)One-hot shape code $\beta$ &26.27$\pm$2.25    &0.895$\pm$0.024    &0.174$\pm$0.039   \\
(a.4)Leanable shape code $\beta$ &25.24$\pm$2.13    &0.883$\pm$0.024    &0.200$\pm$0.041   \\
(a.5){w/o appearance code $\alpha$ \& TEM}    &25.73$\pm$2.03    &0.889$\pm$0.025    &0.184$\pm$0.039   \\

(b.1)Deformation     &24.22$\pm$2.33      & 0.863$\pm$0.027   & 0.231+0.041   \\
(b.2)Modulation   &25.69$\pm$2.22    &0.886$\pm$0.025    &0.187$\pm$0.039   \\
(b.3)Hybrid       &24.42$\pm$2.13      & 0.857$\pm$0.027   & 0.241+0.039   \\

(c)Uniform Sampling   &25.67$\pm$2.09    &0.888$\pm$0.024    &0.185$\pm$0.037   \\
(d)Ours with on changes   &\textbf{26.57$\pm$2.08}    &\textbf{0.897$\pm$0.025}    &\textbf{0.166$\pm$0.037}  \\ \bottomrule
\end{tabular}
\label{tab:ablation}
\vspace{-0.2cm}
\end{table}

\begin{figure}[ht]
    \centering
    \includegraphics[width=1\linewidth, trim=0.1in 0.2in 0.0in 0]{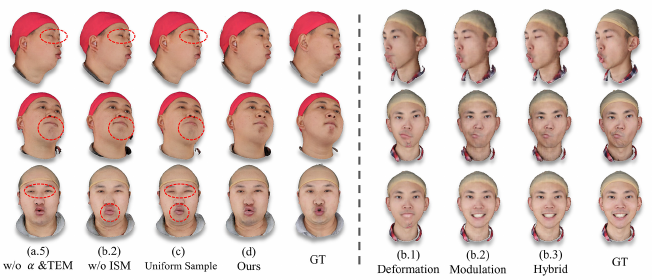}
    \caption{Visual comparison of ablation study. 
    }
    \label{fig:ablation}
    \vspace{-0.5cm}
\end{figure}

\begin{figure}[t]
    \centering
    \includegraphics[width=1.0\linewidth, trim=0 0.05cm 0 0]{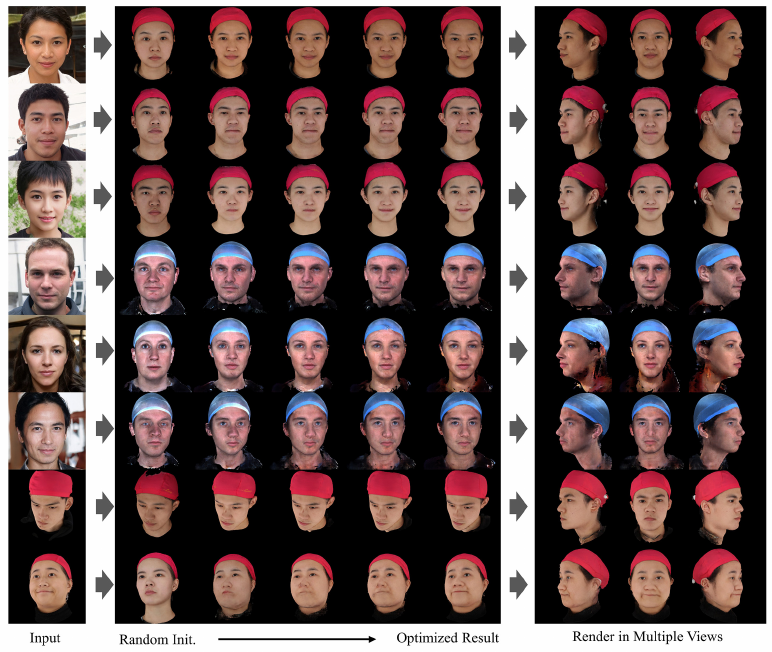}
    \caption{The fitting results to a single-view image of MoFaNeRF. The testing image are from FaceScape testing set and in-the-wild images. The comparison with single-view face reconstruction and failure cases is shown in the supplementary material.
    }
    \vspace{-0.1in}
    \label{fig:fit}
\end{figure}

\begin{figure}[t]
    \centering
    \includegraphics[width=1.0\linewidth]{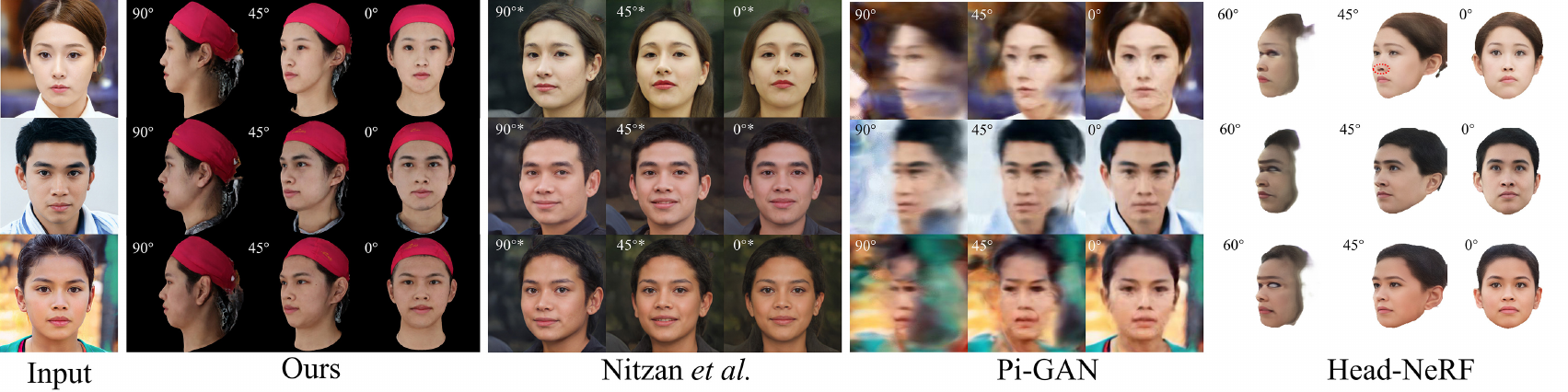}
    \vspace{-0.3in}
    \caption{The fitting and facial rotating results compared with previous methods. 
    }
    \vspace{-0.2in}
    \label{fig:compare_gan}
\end{figure}

\subsection{Ablation Study}
\label{sec:ablation}

We provide ablation studies on coding strategy, morphable NeRF architecture, and sampling strategy:

\noindent$\bullet$ (a) Ablation Study for different coding strategy: one-hot code, PCA code, learnable code. PCA code is the PCA weights generated by the bilinear models\cite{cao2013facewarehouse,yang2020facescape}; Learnable code is optimized in the training of our MoFaNeRF model, which is initialized by a normal distribution. Our method adopts learnable expression code and PCA shape code, so the other 2 choices for expression and shape code are compared. Considering the appearance cannot be coded as one-hot or PCA code, we only compare with and without TEM module and coded appearance in the ablation study. The coding strategy and the TEM module are described in Section~\ref{sec:param} and Section~\ref{sec:network} respectively.

\noindent$\bullet$ (b) Ablation study for morphable NeRF architecture. Previous NeRF variants that support morphable or dynamic objects can be divided into three distinct categories: deformation-based approaches\cite{Tretschk20arxiv_NR-NeRF,park2020deformable,Pumarola20arxiv_D_NeRF}, modulation-based approaches\cite{li2021dynerf,xian2021videoNerf,Li20arxiv_nsff}, and a hybrid of both\cite{park2021hypernerf}. All of these methods were only tested for a single or a small collections, and our ablation study aims to verify their representative ability for a large-scale face dataset. We select NR-NeRF\cite{Tretschk20arxiv_NR-NeRF}, Dy-NeRF\cite{li2021dynerf}, Hyper-NeRF\cite{park2021hypernerf} to represent deformation-based, modulation-based, and hybird architecture respectively. 

\noindent$\bullet$ (c) Ablation study to verify our sampling strategy (Section~\ref{sec:train}). We replace our landmark-based sampling strategy with uniform sampling strategy in the training phase. 

\noindent$\bullet$ (d) Ours with no changes. 

We reconstruct $300$ images of the first $15$ subjects in our training set for evaluation, with random view directions and expressions. The results are reported quantitatively in Table~\ref{tab:ablation} and qualitatively in Figure~\ref{fig:ablation}.

\textbf{Discussion.} As reported in Table~\ref{tab:ablation}, comparing (d) to (a.1) and (a.2), we find the PCA identity code most suitable for encoding shape, which reflects the shape similarity in the parameters space. Comparing (d) to (a.3) and (a.4), we can see that learnable code is most suitable for encoding expressions. 
We think the reason is that the categories of the expression are only $20$, which is quite easy for the network to memorize and parse, while PCA code doesn't help for the few categories. By comparing items (b.1) - (b.3), we can see that modulation-based method (Dy-NeRF) shows better representative ability in modeling large-scale morphable faces, which explain the reason why our final model is based on modulation-based structure. By comparing (b.2) and (d), the positive effect of ISM module explained in Section~\ref{sec:network} is verified. By comparing (c) and (d), we can see our sampling method further boost the performance.
Comparing (a.5), (b.2), (c) to (d), we can see that our proposed TEM, ISM, and landmark-based sampling all have positive effects on model representation ability, and synthesize more faithful results in the visual comparison.

\subsection{Application Results}
\label{sec:Application Result}

\noindent\textbf{Image-based fitting.}
The fitted result to the testing set and in-the-wild images are shown in Figure~\ref{fig:fit}. More results, comparison with single-view reconstruction methods, and failure cases can be found in the supplementary material.

\noindent\textbf{Facial rotating.} 
We fit our model to a single image and rotate the fitted face by rendering it from a side view, as shown in Figure~\ref{fig:compare_gan}. The facial rotating results is compared with Nitzan~\etal~\cite{Nitzan2020FaceID}, Pi-GAN\cite{piGAN2021} and HeadNeRF\cite{hong2022headnerf}. We can see that our method synthesizes a plausible result even at a large angle (close to $\pm90^\circ$) while the facial appearance and shape are maintained. Nitzan~\etal and Pi-GAN are GAN-based networks, while HeadNeRF is a parametric NeRF trained with the help of traditional 3DMM. The results of all these three methods contain obvious artifacts when the face is rotated at a large angle.

\noindent\textbf{Face rigging and editing.}
%\label{sec:manipulate}
As shown in Figure~\ref{fig:title} and Figure~\ref{fig:morph}, after the model is fitted or generated, we can rig the face by driving the expression code $\epsilon$, and edit the face by changing shape code $\beta$ and appearance code $\alpha$. Please watch our results in the video and supplementary materials.

\section{Conclusion}
\label{sec:con}

In this paper, we propose MoFaNeRF that is the first facial parametric model based on neural radiance field.  Different to the previous NeRF variants that focuses on a single or a small collection of objects, our model disentangles the shape, appearance, and expression of human faces to make the face morphable in a large-scale solution space. MoFaNeRF can be used in multiple applications and achieves competitive performance comparing to SOTA methods.

\noindent\textbf{Limitation.} Our model doesn't explicitly generate 3D shapes and focuses on free-view rendering performance. This prevents our model from being directly used in traditional blendshapes-based driving and rendering pipelines. Besides, the single-view fitting of MoFaNeRF only works well for relatively diffused lighting, while the performance will degrade for extreme lighting conditions. In the future work, we believe that introducing illumination model with MoFaNeRF will improve the generalization and further boost the performance. 

\noindent\textbf{Acknowledgement}
This work was supported by the NSFC grant 62025108, 62001213, and Tencent Rhino-Bird Joint Research Program. We thank Dr. Yao Yao for his valuable suggestions and Dr. Yuanxun Lu for proofreading the paper.

\clearpage
% \begin{document}

\section{Supplementary Materials}

\subsection{Overview}
The supplementary material contains a video available at \url{https://neverstopzyy.github.io/mofanerf} and additional descriptions. The video shows a brief overview of our method and the animation of rigging and editing. The additional content contains more results of the image-based fitting (Section~\ref{sec:more_fitting}), failure cases (Section~\ref{sec:fitting_failure}), comparison with previous reconstruction-based view synthesis (Section~\ref{sec:fit_compare}), randomly generated faces (Section~\ref{sec:rand_gen}), the network architecture (Section~\ref{sec:net_param}), details about the training (Section~\ref{sec:train_detail}), and the source of used face images (Section~\ref{sec:data_source}).

\subsection{Animation of Rigging and Editing}
After the face is fitted or generated, it can be rigged by driving the expression code $\epsilon$, and be edited by changing shape code $\beta$ and appearance code $\alpha$. The animation of rigging and editing results are shown in the supplementary video (Part 3 and 4). We can see that the face can be driven by the expression code extracted from a RGB video, and the face can morph smoothly in the dimensions of shape, appearance and expression.

\subsection{More Results of image-based fitting}
\label{sec:more_fitting}
\begin{figure*}
    \centering
    \includegraphics[width=1.0\linewidth]{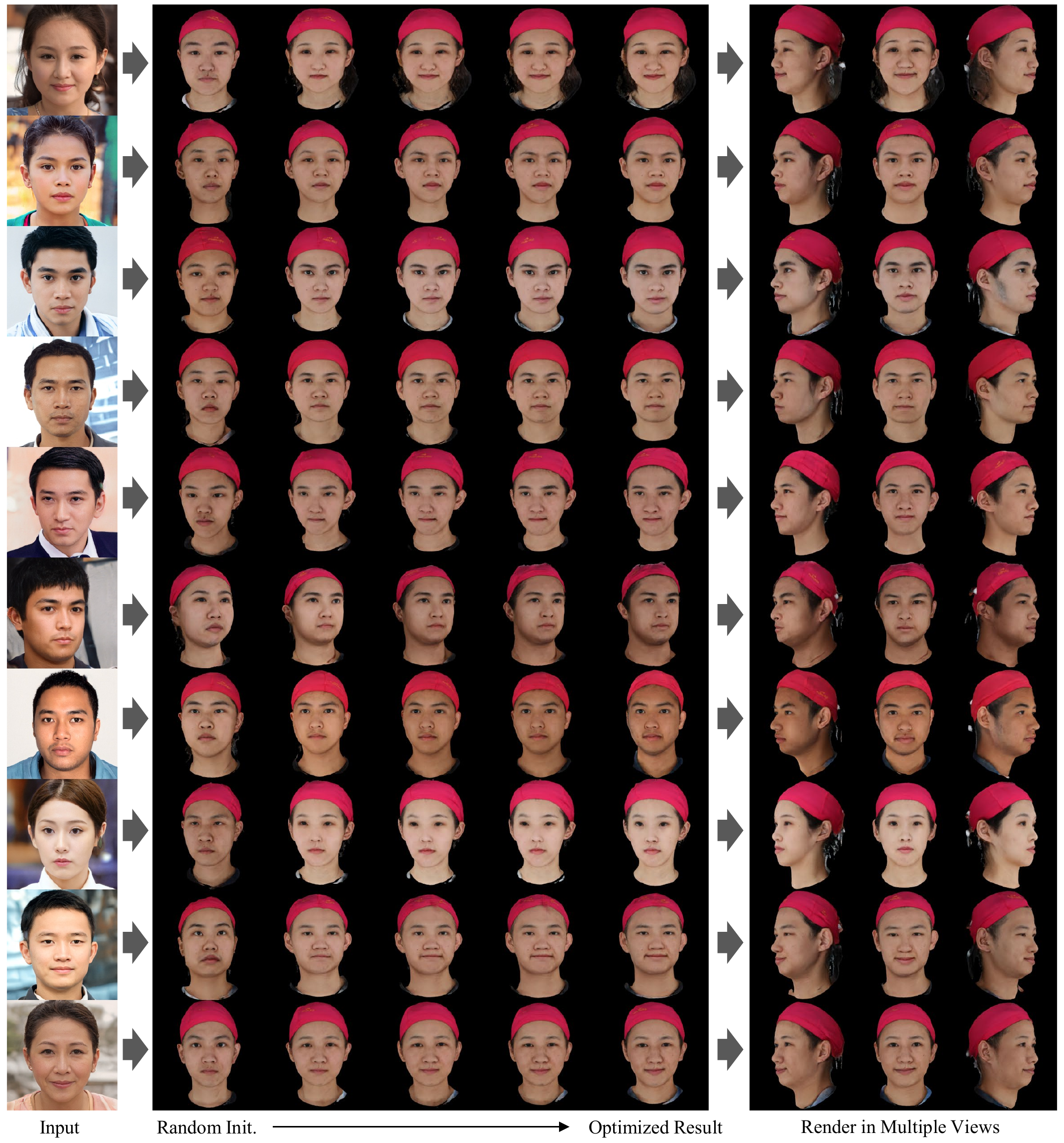}
    \vspace{-0.2in}
    \caption{More fitting results of MoFaNeRF to a single-view image based on FaceScape model.
    }
    \label{fig:more_fit1}
    \vspace{-0.1in}
\end{figure*}

We show more faces fitted to a single image in Figure~\ref{fig:more_fit1} and Figure~\ref{fig:more_fit2}, which is the extension of Figure 7 in the main paper.

\begin{figure}
    \centering
    \includegraphics[width=1.0\linewidth]{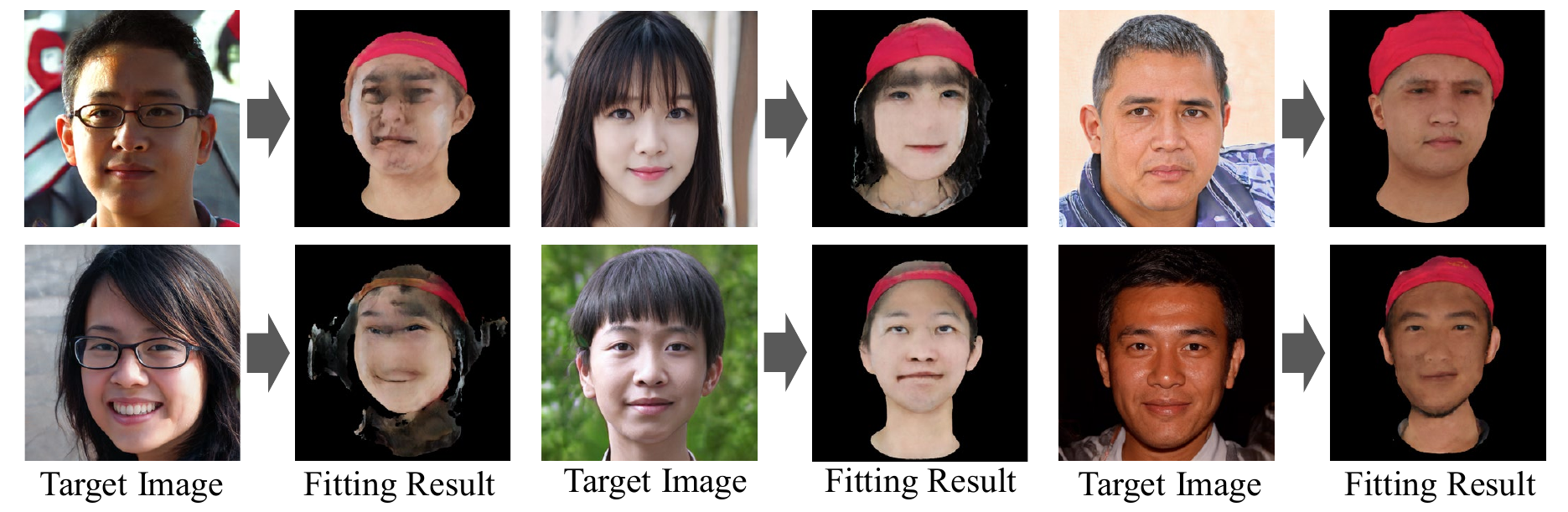}
    \vspace{-0.3in}
    \caption{Failure cases of fitting MoFaNeRF to a single image.}
    \label{fig:failure}
    \vspace{-0.1in}
\end{figure}

\begin{figure*}[t]
    \centering
    \includegraphics[width=1.0\linewidth]{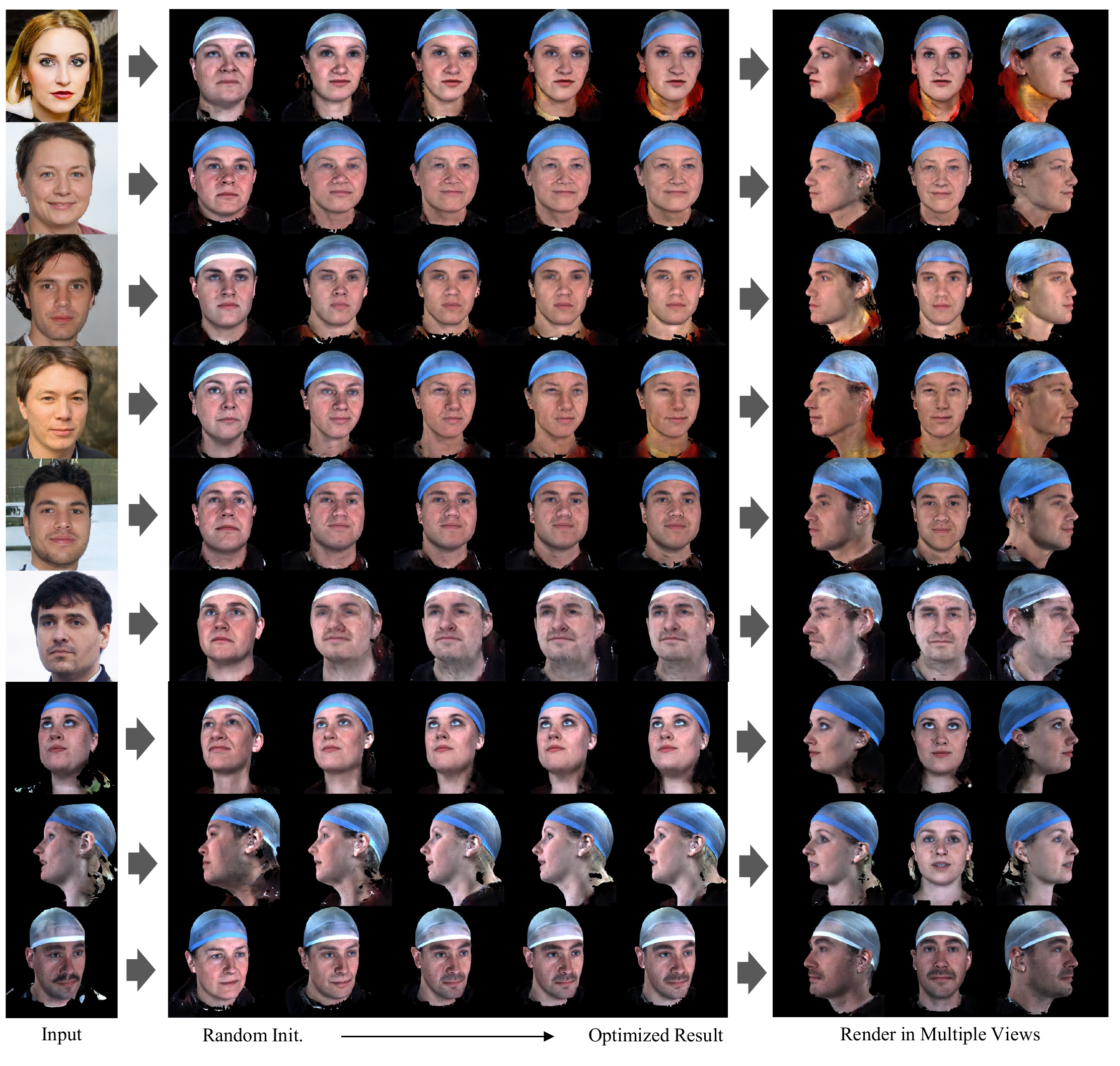}
    \vspace{-0.2in}
    \caption{More fitting results of MoFaNeRF to a single-view image based on HeadSpace model.
    }
    \label{fig:more_fit2}
    \vspace{-0.1in}
\end{figure*}

\subsection{Failure Cases in Fitting}
\label{sec:fitting_failure}

We show some failure cases of the image-based fitting in Figure~\ref{fig:failure}. The first column on the left shows that the fitting results are bad in the extreme lighting. As our model is trained in the images with relatively diffused lighting, large areas of shadow will interfere with our fitting. Lighting models may be introduced in future work to improve the generalization ability for complex lighting conditions. The second column from the left shows that the fitting may fail for the faces with large occluded regions. Fitting MoFaNeRF to an occluded face is still a challenging task to be solved.  The third column from the left shows that the fitting results degraded for the faces that are quite different from the FaceScape dataset in shape (top) or skin color (bottom). The generalization of image-based fitting still needs to be improved.

\begin{figure*}
    \centering
    \includegraphics[width=1.0\linewidth]{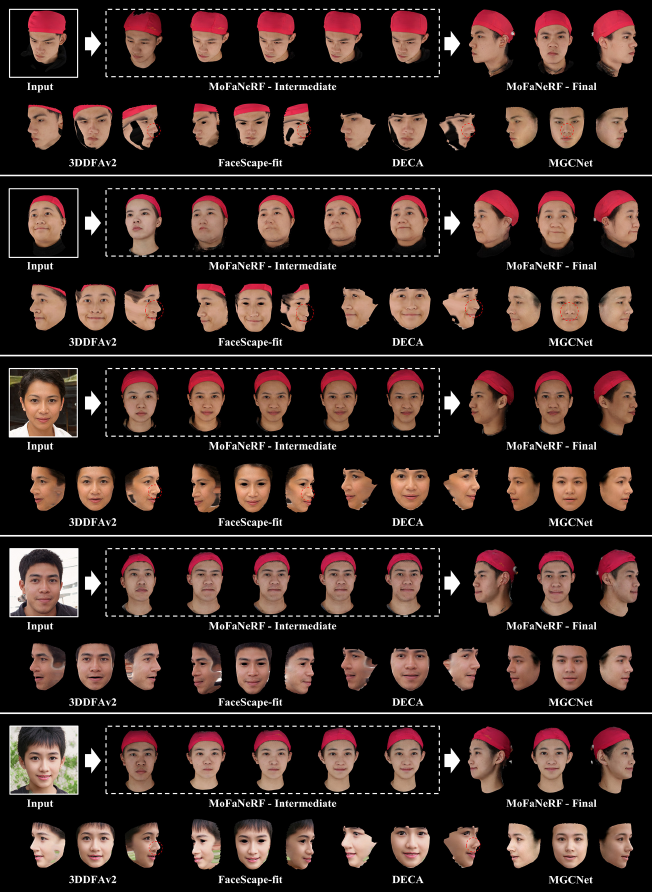}
    \vspace{-0.1in}
    \caption{We compare our fitting and rendering result with SOTA single-view reconstruction methods (SVR). In the red circles, we can see that the inaccurately predicted shape of the nose leads to artifacts in the side view. 3DDFAv2, FaceScape-fit, and DECA commonly contain artifacts on the cheeks due to the misalignment of the predicted shape and the source image. Besides, all four methods cannot align ears well, so no texture is assigned to ears.
    }
    \label{fig:vs_single}
    \vspace{-0.1in}
\end{figure*}

\subsection{Fitting v.s. Single-View Reconstruction}
\label{sec:fit_compare}
As shown in Figure~\ref{fig:vs_single}, we compare our method with four state-of-the-art Single-View Reconstruction(SVR) methods\cite{yang2020facescape,shang2020self,guo2020towards,feng2021learning} in rendering performance. These methods take the single-view image as input and predict the mesh with texture. The images are rendered from the predicted meshes in the frontal view and $\pm60^{\circ}$ side views. Please note that FaceScape-fit\cite{yang2020facescape}, 3DDFAv2\cite{guo2020towards} and DECA\cite{feng2021learning} reconstructed the full head, however, their textures come from the source image and only facial textures are assigned. Therefore, we only render the regions with texture for these three methods.

We can see that the inaccurately predicted shape of FaceScape-fit\cite{yang2020facescape}, 3DDFA-v2\cite{guo2020towards} and DECA\cite{feng2021learning} leads to the artifacts in the side views, as shown in the red dotted circles. Besides, these methods commonly contain wrong scratches on the cheeks due to the misalignment of the predicted shape and the source image. Though the MGCNet doesn't contain the scratches problems, its texture tends to be a mean texture with less detail. We can also observe that in some cases the shape of the nose is unfaithful. Besides, all four methods cannot align ears well, so no texture is assigned to ears. By contrast, our rendering results contain fewer artifacts and are more plausible in the side views.  The ears are also rendered in our method, which makes the side-view rendering complete.

\begin{figure*}[t]
    \centering
    \includegraphics[width=1.0\linewidth]{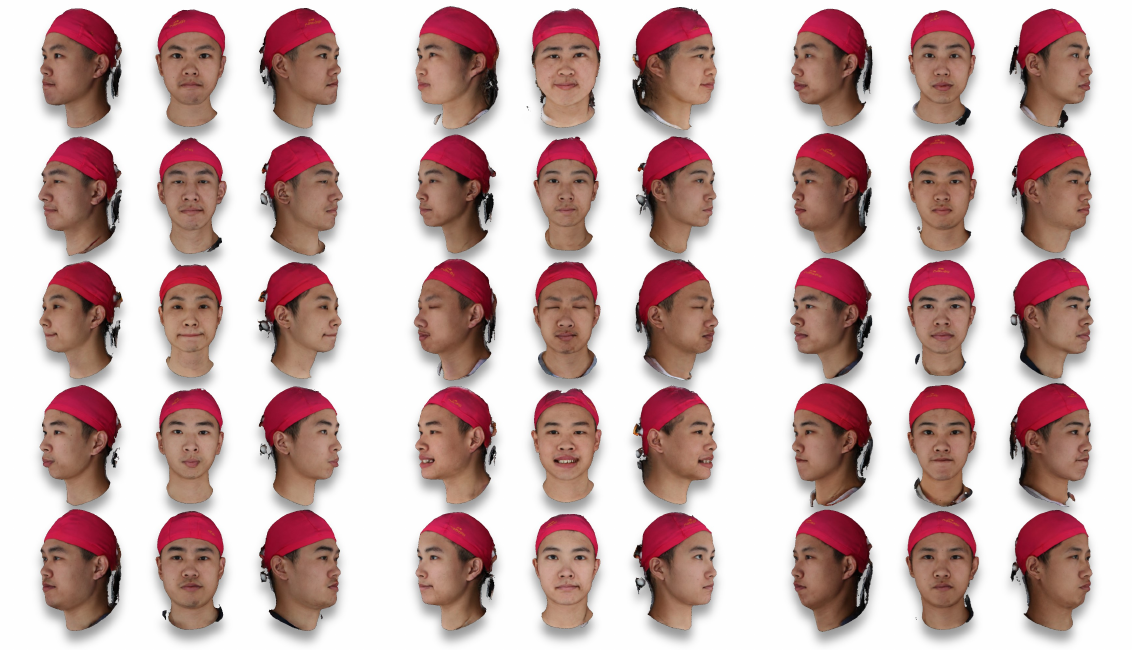}
    \vspace{-0.3in}
    \caption{More random generated results by our models. We visualize them in three views with yaw angles in $[-60^{\circ},0^{\circ},60^{\circ}]$ and pitch angles in $0^{\circ}$.
    }
    \label{fig:more_gen}
    \vspace{-0.1in}
\end{figure*}

\subsection{Results of Random Generation}
\label{sec:rand_gen}
Some randomly generated faces are shown in Figure~\ref{fig:more_gen}. We can see that our model covers a wide range of facial shapes, appearances and expressions.

\subsection{Disentanglement between appearance and geometry}
\begin{figure}[h]
    \centering
    \includegraphics[width=1.0\linewidth]{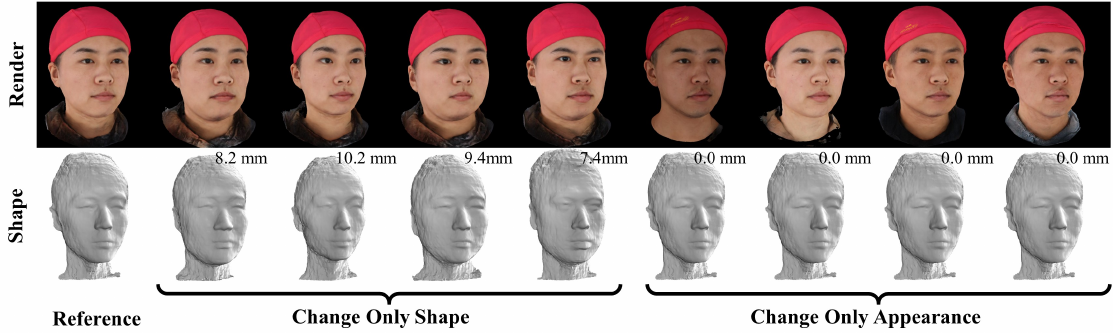}
    \vspace{-0.3in}
    \caption{Extracted shape for verifying disentanglement.} 
    \label{fig:disen_app_and_geo}
    \vspace{-0.1in}
\end{figure}

In addition to Fig. 3 of the main paper, we supplement the experiment in Fig.~\ref{fig:disen_app_and_geo} to extract the shape from the radiance fields, and report the chamfer distance to quantitatively measure the shape consistency. The figure and quantitative results show that the extracted shape is changed only when the shape code is changed, which indicates the effective disentanglement of shape from appearance and expression.
\label{sec:disen_app_and_geo}

\subsection{Parameters of Network}
\label{sec:net_param}

\begin{figure}[t]
    \centering
    \includegraphics[width=0.75\linewidth, scale=0.8]{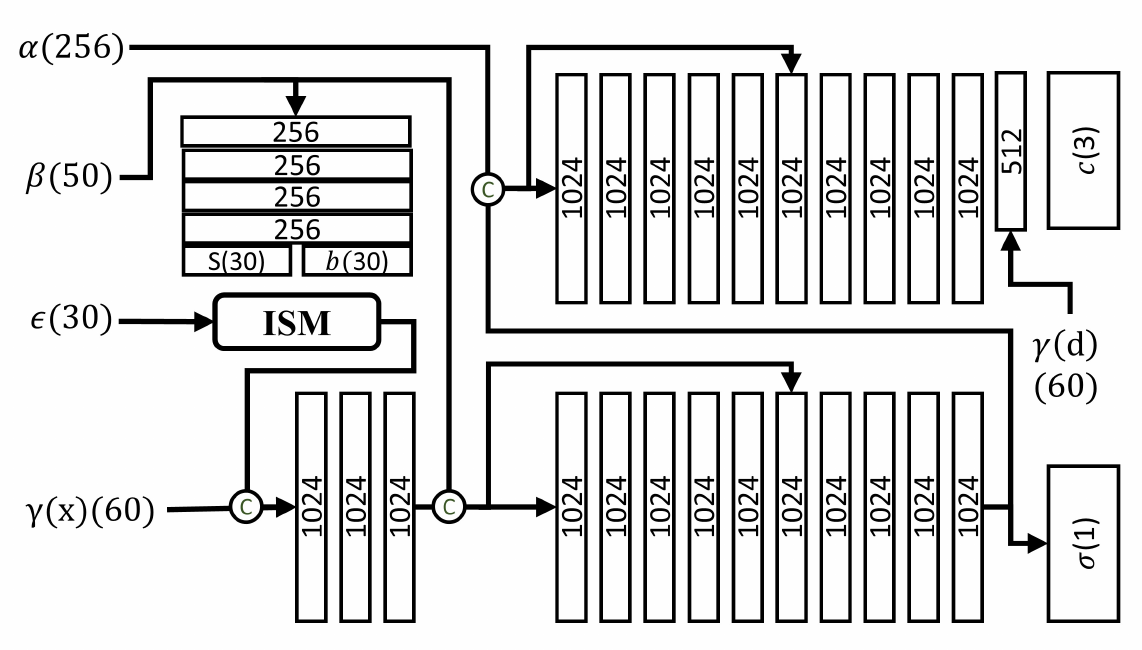}
    \vspace{-0.2in}
    \caption{The detailed parameters of the network in MoFaNeRF. The number in brackets indicate the length of the tensor.}
    \label{fig:network_par}
    % \vspace{-0.5in}
\end{figure}

The parameters of our network are shown in Figure~\ref{fig:network_par}. The boxes represent the full connection layer, where the numbers represent the number of neurons. The circles with C inside represent the concatenating operation between tensors. The numbers follow the parameter means the dimension of this parameter.  $\alpha, \beta, \epsilon$ are the parameters of appearance, shape and expression. $\gamma(\textbf{x})$ and $\gamma(\textbf{d})$ mean the position encoding of the position code $\textbf{x}$ and viewing direction $\textbf{d}$. $\sigma$ and $\textbf{c}$ are the density and color that form the radiance field. The parameters of the TEM module are shown in Table~\ref{tab:tem}.

\begin{table}[t]
\centering\caption{The detailed parameters of the TEM in MoFaNeRF.
All convolution layers and linear layers are followed by Leaky ReLU\cite{maas2013rectifier} with negative slope of 0.2 and 0.1 respectively, except for layers ``mu" and ``std". 'Repara.': means the reparameterization method\cite{kingma2013auto,wang2021learning} to produce latent code from the distribution $\mathcal{N}(\mathbf{\mu},\mathbf{\sigma})$. $k$: kernel size ($k \times k$).  $s$: stride in both horizontal and vertical directions.  $p$: padding size ($p \times p$).  $c$: number of output channels.  $d$: output spatial dimension ($d \times d$).  `Conv': convolution layer.  `Linear': fully connected layer. 'Flatten': flatten layer.}
\begin{tabular}{ccccccc}\toprule
Name   & Type         & input       & (k,s,p) & c    & d   \\ \midrule
conv1  & Conv         & textureMap  & (4,2,1) & 32   & 256 \\
conv2  & Conv         & conv1       & (4,2,1) & 32   & 128 \\
conv3  & Conv         & conv2       & (4,2,1) & 32   & 64  \\
conv4  & Conv         & conv3       & (4,2,1) & 32   & 32  \\
conv5  & Conv         & conv4       & (4,2,1) & 64   & 16  \\
conv6  & Conv         & conv5       & (4,2,1) & 128  & 8   \\
conv7  & Conv         & conv6       & (4,2,1) & 256  & 4   \\
flat0  & Flatten      & conv7       &\ \,--$^*$        & 4096 & 1    \\
line1  & Linear       & flat0       &--         & 512  & 1    \\
mu     & Linear       & line1       &--   & 256  & 1    \\
logstd & Linear       & line1       &--       & 256  & 1    \\
para.  & Repara. & (mu,logstd) &--        & 256  & 1    \\
line2  & Linear       & para.       &--       & 256  & 1    \\
line3  & Linear       & line2       &--        & 256  & 1    \\
app.  & Linear       & line3       &--       & 256  & 1    \\\bottomrule
\end{tabular}
\leftline{ \small{* `--' means meaningless parameters.}}
\label{tab:tem}
\end{table}\textbf{}

\subsection{Details of Training and Data}
\label{sec:train_detail}

\noindent\textbf{Implement details.} Our model is implemented on PyTorch~\cite{PyTorch}. In our experiment, $1024$ rays are sampled in an iteration, each with 64 sampled points in the coarse volume and additional $64$ in the fine volume.  The strategy of hierarchical volume sampling is used to refine coarse volume and fine volume simultaneously, as defined in NeRF\cite{mildenhall2020nerf}. The resolution of the images rendered by MoFaNeRF is $256\time256$, and the RefineNet takes the image rescaled to $512\times512$ as input, and synthesizes the final image in $512\times512$.

We use the Adam optimizer\cite{kingma2015adam} with the initial learning rate as $5\times{10^{-4}}$, decayed exponentially as $2\times{10^{-5}}$, $\beta_1=0.9$, $\beta_2=0.999$,  $\epsilon=10^{-7}$. Our model is trained for roughly $400k$ iterations, which takes about 2 days on dual NVIDIA GTX3090 GPUs.

\begin{figure*}[t]
    \centering
    \includegraphics[width=1.0\linewidth]{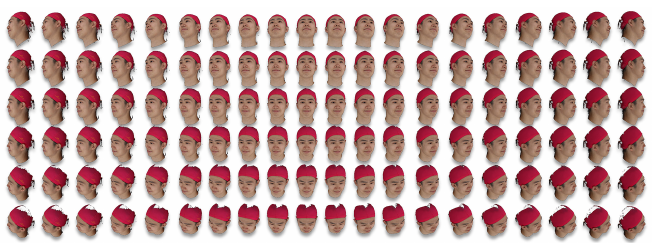}
    \vspace{-0.3in}
    \caption{Multi-view images are generated by FaceScape in $120$ views with $6$ pitch angles in $[-30^\circ\sim+45^\circ]$ and $20$ yaw angles in $[-90^\circ\sim+90^\circ]$.
    }
    \label{fig:rend_view1}
    \vspace{-0.1in}
\end{figure*}

\begin{figure*}[t]
    \centering
    \includegraphics[width=1.0\linewidth]{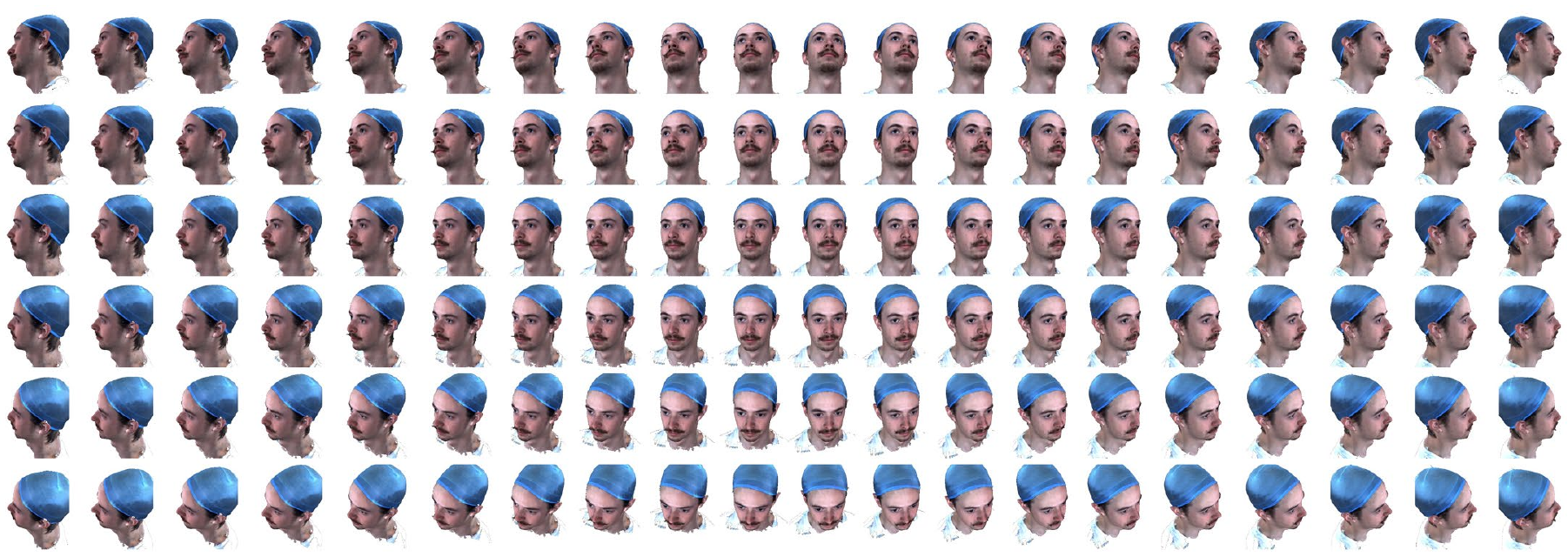}
    \vspace{-0.3in}
    \caption{Multi-view images are generated by HeadSpace in $120$ views with $6$ pitch angles in $[-30^\circ\sim+45^\circ]$ and $20$ yaw angles in $[-90^\circ\sim+90^\circ]$.
    }
    \label{fig:rend_view2}
    \vspace{-0.1in}
\end{figure*}

\noindent\textbf{Details about data preparation.} Taking face orientation as the reference direction, we evenly select $6$ pitch angles from $-30^\circ$ to $45^\circ$ and $20$ yaw angles from $-90^\circ$ to $+90^\circ$ degrees for rendering in $120$ viewpoints. The samples of $120$ views are shown in Figure~\ref{fig:rend_view1} and Figure~\ref{fig:rend_view2}. We use all the training data to train the network of MoFaNeRF, and randomly select $24,000$ rendering results to train the RefineNet.

Initially, we plan to use the raw scanned multi-view images released by FaceScape~\cite{yang2020facescape,zhu2021facescape}, however, we find the camera locations are not uniform for all these $7180$ tuples of images. We contacted the authors of FaceScape and learned that the reason was that the capturing took place in two locations, where the camera setups and parameters were changed several times.  Therefore, we use the multi-view images to color the raw scanned models, then render them according to the viewpoints as set above to obtain high-fidelity multi-view images with uniform camera parameters.

\subsection{Face Image Source}
\label{sec:data_source}
To avoid portrait infringement, we used some synthesized `in-the-wild' face images for testing our model. The source images in Figure~{\color{red}1, 4, 9, 10} are synthesized by StyleGANv2\cite{karras2020analyzing}, which is released under Nvidia source code license. So the use of these virtual portraits will not raise infringement issues. We have signed the license agreement with the authors of FaceScape~\cite{yang2020facescape,zhu2021facescape} to obtain the permission to use the dataset for non-commercial research purpose, and the permission to publish the subjects of $12$, $17$, $40$, $49$, $57$, $92$, $97$, $168$, $211$, $212$, $215$, $234$, $260$, $271$, $326$ in Figure~{\color{red}2, 3, 5, 6, 7, 8} of this paper.

\clearpage
\bibliographystyle{splncs04}
\bibliography{mybib}

\end{document}